# Es igual pero no es lo mismo: ¿Distinguen los LLMs las variedades del español?

*It's the same but not the same: Do LLMs distinguish Spanish varieties?*

**Marina Mayor-Rocher[1], Cristina Pozo[2], Nina Melero[2,3], Gonzalo Martínez[4], María Grandury[2,5], Pedro Reviriego[2]**

[1] Universidad Autónoma de Madrid
[2] Universidad Politécnica de Madrid
[3] New York University
[4] Universidad Carlos III de Madrid
[5] SomosNLP

Contacto: marina.mayor@uam.es

**Resumen:** En los últimos años, los grandes modelos de lenguaje (LLMs, por sus siglas en inglés) han demostrado una alta capacidad para comprender y generar texto en español. Sin embargo, con quinientos millones de hablantes nativos, la española no es una lengua homogénea, sino rica en variedades diatópicas que se extienden a ambos lados del Atlántico. Por todo ello, evaluamos en este trabajo la capacidad de nueve modelos de lenguaje de identificar y discernir las peculiaridades morfosintácticas y léxicas de siete variedades de español (andino, antillano, caribeño continental, chileno, español peninsular, mexicano y centroamericano y rioplatense) mediante un test de respuesta múltiple. Los resultados obtenidos indican que la variedad de español peninsular es la mejor identificada por todos los modelos y que, de entre todos, GPT-4o es el único modelo capaz de identificar la variabilidad de la lengua española.

**Palabras clave:** Grandes modelos de lenguaje, variedades del español, evaluación, dialectología.

**Abstract:** In recent years, large language models (LLMs) have demonstrated a high capacity for understanding and generating text in Spanish. However, with five hundred million native speakers, Spanish is not a homogeneous language but rather one rich in diatopic variations spanning both sides of the Atlantic. For this reason, in this study, we evaluate the ability of nine language models to identify and distinguish the morphosyntactic and lexical peculiarities of seven varieties of Spanish (Andean, Antillean, Continental Caribbean, Chilean, Peninsular, Mexican and Central American and Rioplatense) through a multiple-choice test. The results indicate that the Peninsular Spanish variety is the best identified by all models and that, among them, GPT-4o is the only model capable of recognizing the variability of the Spanish language.

## 1 Introducción

Los grandes modelos de lenguaje (LLMs) han supuesto un avance sin precedentes en el procesamiento del lenguaje natural, si bien su desarrollo ha estado inicialmente dominado por el inglés como lengua base de entrenamiento. En efecto, aunque el español cuenta con 600 millones de hispanohablantes alrededor del mundo, su presencia en los conjuntos de datos de entrenamiento sigue siendo escasa. Como señalan Muñoz-Basols et al. (2024), este predominio del inglés en el entrenamiento de los modelos genera lo que han denominado el Sesgo Lingüístico Digital (SLD), tanto a nivel interlingüístico, por la influencia de las estructuras y el léxico anglófono, como a nivel intralingüístico, pues se priorizan determinadas variedades del español en detrimento de otras.

Ejemplos de ello los encontramos en modelos del proyecto MarIA como RoBERTa y GPT-2, cuyos datos de entrenamiento, además de los del inglés, se obtuvieron de la Biblioteca Nacional Española, representando en un 50% a la variedad española peninsular (Muñoz-Basols et al. 2024: 631). Lo mismo ocurre con el proyecto CEREAL (Corpus del Español REAL), dedicado a compilar y a anotar documentos en español recopilados de la web, puesto que la mayoría de sus datos, extraídos de OSCAR (Open Super-large Crawled ALMAnaCH Repository), representaban la variedad de español peninsular (España-Bonet y Barrón-Cedeño 2024: 3691).

Esta desproporción de datos no solo afecta a la capacidad del modelo para interpretar y generar nuevos textos y datos redactados correctamente en español, sino que invisibiliza al mismo tiempo las peculiaridades lingüísticas o dialectales de comunidades de habla enteras. Para evitar que los LLMs perpetúen un español homogeneizado y una desigualdad y discriminación lingüísticas que ya existen en el imaginario de muchos hablantes reales, es esencial detectarlas eficazmente mediante el desarrollo de metodologías de evaluación que midan con precisión su desempeño.

Así, el propósito de este trabajo es el de evaluar qué capacidad tienen nueve modelos de lenguaje para diferenciar siete variedades del español: andina, antillana, caribeña continental, chilena, español peninsular, mexicana y centroamericana y rioplatense. Los objetivos principales son dos: saber, por un lado, qué variedades de español son las mejor reconocidas por los modelos y, por el otro, qué modelos realizan mejor esta tarea. Solo mediante este ejercicio de evaluación podrá entenderse el correcto o incorrecto funcionamiento de los modelos en lengua española y en sus diferentes variedades para determinar, al mismo tiempo, si los procesos de entrenamiento de los modelos son justos y representativos. A esta introducción le siguen la metodología, el análisis de los resultados obtenidos para cada variedad, la discusión de los resultados, las conclusiones y líneas futuras y la bibliografía citada.

## 2  Metodología

El objetivo principal de este estudio es analizar el conocimiento de las diferentes variedades de español que tienen los LLMs. Para ello, se consideran las variedades de la Tabla 1, basadas en la clasificación convencional de la dialectología hispánica (Gutiérrez-Rexach, 2016; Moreno Fernández, 2014), que son también las mismas áreas que emplea ASALE para distribuir geográficamente sus bases de datos, el CREA y el CORPES XXI.

| Variedad |
|---|
| Andino |
| Antillano |
| Caribeño continental |
| Chileno |
| Español peninsular |
| México y Centroamérica |
| Rioplatense |

Tabla 1: Variedades analizadas

Conscientes de las limitaciones de esta división, las variedades han sido normalizadas para una primera prueba, puesto que el espectro de variedades de español es demasiado amplio. Así, para el diseño de las preguntas y respuestas, se ha seleccionado la variedad estándar de cada una de ellas. Por ejemplo, para la variedad estándar peninsular, se tiene en cuenta la septentrional y se excluyen otras.

Para evaluar a los distintos LLMs se ha diseñado una prueba con 30 preguntas de opción múltiple, donde cada respuesta está asociada a una o varias variedades del español. Se compone de 20 preguntas que indagan sobre variación morfosintáctica —por ejemplo, el uso de formas verbales en diferentes contextos, el orden de los sintagmas en la oración o los sistemas pronominales—, así como 10 preguntas sobre variación léxica —denominaciones de objetos de uso cotidiano o léxico de registro coloquial—. Para asegurar que el cuestionario fuera representativo de todas las variedades, las distintas preguntas de morfosintaxis se formularon tomando de base al menos un fenómeno gramatical característico de cada una de las variedades, de acuerdo a aquellos descritos en manuales de lingüística hispánica (Gutiérrez-Rexach, 2016; Alvar, 1996; Bosque et al. 1999) y diversos estudios más focalizados en cada variedad. Por otro lado, las preguntas de léxico se basaron en el *Diccionario de la lengua española* (RAE, 2024) y el *Diccionario de americanismos* (ASALE, 2010), así como el proyecto VARILEX (Ueda, 2016). Dado que todas las preguntas son exactamente las mismas para cada una de las variedades, aunque haya preguntas enfocadas en fenómenos específicos, las respuestas posibles contienen opciones válidas para todas ellas. Asimismo, la autenticidad y frecuencia de los fenómenos representados en las preguntas y sus respuestas correspondientes se confirmaron a través de varios bancos de datos del español, como CORPES XXI, CREA, CORDE, Web/Dialects y el corpus del PREESEA. De este modo, se comprueba que las opciones aquí seleccionadas tienen representatividad real en cada una de las variedades y en Internet.

El procedimiento consiste en aplicar esta serie de preguntas de opción múltiple para evaluar cómo responde el LLM en diferentes contextos lingüísticos, según un rol. Es decir, se trata de un examen sobre las distintas variedades del español a modelos que han de responder como hablantes nativos de esa variedad. De esta manera, los modelos han de seleccionar la respuesta correcta para cada pregunta en cada variedad. A continuación, se presentan un par de ejemplos de preguntas del test con su rol especificado y las respuestas esperadas:

*«Eres de Centroamérica, nacido en México, Guatemala, Costa Rica, Honduras, Nicaragua, Panamá o El Salvador. Responde a la siguiente pregunta con la opción que te resulte más natural. Ajústate solo a las opciones dadas.»*

**P. 8: *¿Cuál suena más natural?***

a. «Llegas tarde, vístete y corre».
b. «Llegas tarde, vístete y córrele».

Donde 'a' sería la respuesta elegida naturalmente para español peninsular, andino, antillano, caribeño continental, chileno y rioplatense, mientras que 'b' para mexicano y centroamericano.

**P. 28: *¿Qué verbo usas para describir la acción de ponerse de pie?***

*a. levantarse*
*b. pararse*

Donde 'a' sería la respuesta elegida naturalmente para español peninsular y chileno, mientras que 'b' para andino, antillano, caribeño continental, mexicano y centroamericano, chileno y rioplatense.

Para evaluar las respuestas del modelo, se han utilizado las estimaciones de probabilidad del propio modelo para cada opción para determinar la respuesta seleccionada. Es decir, seleccionamos la opción con la mayor probabilidad estimada por el modelo (log-prob) como su elección final. Las respuestas correctas recibieron una puntuación de 1, mientras que las incorrectas se penalizaron con la siguiente fórmula:

$$Penalización = -\frac{Respuestas\ correctas\ posibles}{Número\ de\ opciones}$$

De este modo se penalizan más los errores en preguntas más sencillas, donde acertar sería más probable, y menos en aquellas en las que la elección es más difícil.

Los modelos empleados en los experimentos se recogen en la Tabla 2. En total, se evaluaron nueve modelos. Siete de ellos son modelos de pesos abiertos y dos son modelos propietarios. La selección incluye modelos de grandes empresas como Google (Riviere et al. 2024), Meta (Grattafiori et al. 2024) y OpenAI (GPT4), así como modelos desarrollados por empresas más pequeñas como Mistral (Jiang et al. 2023) o centrados en idiomas distintos del inglés como Yi y Occiglot (Avramidis et al. 2024). En conjunto, constituyen una muestra representativa de los LLM disponibles a principios del 2025.

Los modelos GPT se utilizaron mediante su API oficial, mientras que el resto de los modelos se ejecutó en local, tras la descarga de los pesos en Hugging Face. Los hiperparámetros del modelo han sido los predeterminados de la plataforma, con la excepción de la temperatura, que es el parámetro que controla la aleatoriedad de las respuestas y se ha fijado en 0 para garantizar resultados deterministas en los que se selecciona siempre la opción con mayor probabilidad estimada por el modelo.

| Modelo |
|---|
| GPT-4o |
| GPT-4o-mini |
| Gemma-2-9b-it |
| Mistral-7B-Instruct-v0.3 |
| Llama-3.2-11B-Vision-Instruct |
| Llama-3.1-8B-Instruct |
| Llama-3.2-3B-Instruct |
| Occiglot-7b-es-en-instruct |
| Yi-1.5-9B-Chat |

Tabla 2: Modelos evaluados

## 3  Análisis

Los resultados de todos los modelos para las 30 preguntas están disponibles en un repositorio público para su estudio[1]. A continuación, presentamos una breve discusión de los resultados y un resumen estadístico descriptivo de los mismos para cada variedad del español.

En las tablas 3 a 9, se proporcionan tres tipos de datos. Por una parte, la puntuación que obtuvo cada modelo según la fórmula explicada previamente; segundo, el índice absoluto de aciertos; y, por último, los porcentajes de respuestas erróneas según la tipificación de las preguntas (morfosintácticas y léxicas), relativos al número de preguntas de cada tipo. Así, se puede comprobar si los modelos son más competentes al identificar cuestiones de variación geolectal de ámbito léxico o gramatical.

### 3.1  Andino

La mayoría de modelos no son capaces de reconocer los fenómenos particulares de la variedad andina, como se ve en la Tabla 3. El modelo con mejor rendimiento es GPT-4o; del resto, solo Llama-3.2-8B-instruct supera un índice de acierto por encima de diez puntos. En general, las preguntas de léxico son relativamente las peores, con la excepción de los modelos de GPT-4o, que porcentualmente tienen la misma cantidad de errores en léxico y en morfosintaxis.

De las preguntas gramaticales, solo Occiglot eligió las variantes de loísmo andino y de supresión del pronombre acusativo; así, por ejemplo, el 77% de los modelos prefirieron *Fui a ver la carretera y ya la habían arreglado* (P. 3) y *Esos bultos vas a llevarlos a la tienda* (P. 2), que son las variantes generales, en lugar de *Fui a ver la carretera y ya lo habían arreglado* o *Esos bultos vas a llevar a la tienda*, ejemplos reales de hablantes andinos (Fernández Ordóñez, 1999). De léxico, la pregunta que todos los modelos fallaron fue la 25 («¿Qué palabra usas para nombrar a la legumbre que es semilla de la planta *phaseolus vulgaris*?»), en la que ninguno de los modelos fue capaz de identificar la palabra más frecuente en esta región: «habas», según datos del corpus Web/Dialects (Davies, n.d.).

| Modelos | Puntuación | Aciertos | Frecuencia relativa de fallos |
|---|---|---|---|
| GPT-4o | 17.6/30 | 21/30 | 30% Morfosintaxis 30% Léxico |
| GPT-4o-mini | 7.1/30 | 14/30 | 55% Morfosintaxis 50% Léxico |
| Gemma-2-9b-it | 5.2/30 | 12/30 | 45% Morfosintaxis 70% Léxico |
| Mistral-7b | 5.4/30 | 12/30 | 40% Morfosintaxis 100% Léxico |
| Llama-3.2-11B-Vision-Instruct | 5.7/30 | 12/30 | 50% Morfosintaxis 80% Léxico |
| Occiglot-7b-es-en-instruct | 8.7/30 | 14/30 | 35% Morfosintaxis 90% Léxico |
| Llama-3.1-8B-Instruct | 5.9/30 | 12/30 | 50% Morfosintaxis 80% Léxico |
| Llama-3.2-3B-Instruct | 10.2/30 | 12/30 | 50% Morfosintaxis 80% Léxico |
| Yi-1.5-9B-Chat | 8/30 | 14/30 | 40% Morfosintaxis 80% Léxico |

Tabla 3: Relación de puntuación, aciertos y fallos de los modelos para español andino.

### 3.2  Antillano

Para la variedad de español antillano, los modelos fallan fundamentalmente las preguntas relacionadas con el léxico cotidiano (prendas de ropa, verdura, objetos) como se ve en la Tabla 4; y encuentran también dificultades para responder a las preguntas que examinan el uso y posición de los pronombres. La mayoría de los modelos no detectan las diferencias entre formulaciones que presentan un uso regional característico, como la aparición del sujeto antepuesto al infinitivo y la ausencia de inversión del sujeto en las preguntas (Morales, 1999). Algunos ejemplos de esas preguntas en las que los modelos han fallado son: «¿Cómo tú te llamas?» y «Lo consultaré para yo saberlo».

| Modelos | Puntuación | Aciertos | Frecuencia relativa de fallos |
|---|---|---|---|
| GPT-4º | 18.2/30 | 21/30 | 30% Morfosintaxis 30% Léxico |
| GPT-4o-mini | 13.9/30 | 18/30 | 40% Morfosintaxis 40% Léxico |
| Gemma-2-9b-it | 8.3/30 | 14/30 | 55% Morfosintaxis 50% Léxico |
| Mistral-7b | 4.5/30 | 11/30 | 65% Morfosintaxis 80% Léxico |
| Llama-3.2-11B-Vision-Instruct | 2.6/30 | 10/30 | 45% Morfosintaxis 70% Léxico |
| Occiglot-7b-es-en-instruct | 8.2/30 | 14/30 | 45% Morfosintaxis 70% Léxico |
| Llama-3.1-8B-Instruct | 5.3/30 | 12/30 | 55% Morfosintaxis 70% Léxico |

---

[1] https://zenodo.org/records/15101403

| | 12.35/30 | 10/30 | 65% Morfosintaxis |
| --- | --- | --- | --- |
| Llama-3.2-3B-Instruct | | | 70% Léxico |
| Yi-1.5-9B-Chat | 5.3/30 | 12/30 | 55% Morfosintaxis 50% Léxico |

Tabla 4: Relación de puntuación, aciertos y fallos de los modelos para español antillano.

### 3.3 Caribeño continental

Los modelos no parecen detectar las peculiaridades de esta variedad de español, sobre todo en el plano gramatical, en preguntas que indagan en el orden sintáctico de los pronombres, como en la pregunta 17, para la cual encuentran naturales las respuestas incorrectas «Vos ya sabés lo que quiero decir» o «Tú ya sabes lo que quiero decir» en vez de la solución con el pronombre pospuesto: «Ya tú sabes lo que quiero decir». En general, los modelos seleccionan el pronombre *le* para el objeto directo de persona en esta variedad, que suele optar por la forma etimológica (Aleza Izquierdo 2010: 113) (P. 14 «¿Le has llamado?»). Asimismo, para esta variedad, los modelos no son capaces de identificar las peculiaridades léxicas que la caracterizan (responde *frijoles* o *alubias* en vez de *caraotas* en la pregunta 25 «¿Qué palabra usas para nombrar a la legumbre que es semilla de la planta *phaseolus vulgaris*?») y los fallos en morfosintaxis no bajan del 25% en el mejor modelo, como se observa en la Tabla 5.

| Modelos | Puntuación | Aciertos | Frecuencia relativa de fallos |
| --- | --- | --- | --- |
| GPT-4o | 19.3/30 | 22/30 | 25% Morfosintaxis 30% Léxico |
| GPT-4o-mini | 13.2/30 | 18/30 | 45% Morfosintaxis 30% Léxico |
| Gemma-2-9b-it | 13.5/30 | 18/30 | 40% Morfosintaxis 40% Léxico |
| Mistral-7b | 8/30 | 16/30 | 45% Morfosintaxis 50% Léxico |
| Llama-3.2-11B-Vision-Instruct | 6.5/30 | 14/30 | 50% Morfosintaxis 60% Léxico |
| Occiglot-7b-es-en-instruct | 7.9/30 | 14/30 | 45% Morfosintaxis 70% Léxico |
| Llama-3.1-8B-Instruct | 4.7/30 | 13/30 | 55% Morfosintaxis 60% Léxico |
| Llama-3.2-3B-Instruct | 10.3/30 | 12/30 | 55% Morfosintaxis 70% Léxico |
| Yi-1.5-9B-Chat | 5.4/30 | 12/30 | 50% Morfosintaxis 80% Léxico |

Tabla 5: Relación de puntuación, aciertos y fallos de los modelos para caribeño continental.

### 3.4 Chileno

La mayor parte de los modelos no detectan las elecciones léxicas a las que la variedad chilena da preferencia (adjetivos y nombres comunes; por ejemplo, «palabra para referirse a un amigo de forma familiar» o «nombre para designar el tubo por el que se sorbe una bebida»; etc.; de acuerdo con las entradas clasificadas como chilenismos («weón», «bombilla») en el diccionario de la RAE [2014]). Por otro lado, no distinguen marcas regionales en el uso de los pronombres. Para esta variedad en concreto, hay que señalar que GPT-4o es el único modelo que consigue casi un 100% de los aciertos en cuanto a léxico, con un único error de este tipo. Los resultados de los diferentes modelos se resumen en la Tabla 6.

| Modelos | Puntuación | Aciertos | Frecuencia relativa de fallos |
| --- | --- | --- | --- |
| GPT-4o | 20.4/30 | 24/30 | 25% Morfosintaxis 10% Léxico |
| GPT-4o-mini | 16.9/30 | 21/30 | 35% Morfosintaxis 20% Léxico |
| Gemma-2-9b-it | 12.9/30 | 18/30 | 30% Morfosintaxis 60% Léxico |
| Mistral-7b | 10.2/30 | 17/30 | 30% Morfosintaxis 70% Léxico |
| Llama-3.2-11B-Vision-Instruct | 7.5/30 | 15/30 | 35% Morfosintaxis 80% Léxico |
| Occiglot-7b-es-en-instruct | 12.8/30 | 15/30 | 40% Morfosintaxis 70% Léxico |
| Llama-3.1-8B-Instruct | 5.8/30 | 14/30 | 40% Morfosintaxis 80% Léxico |
| Llama-3.2-3B-Instruct | 10.2/30 | 13/30 | 45% Morfosintaxis 80% Léxico |
| Yi-1.5-9B-Chat | 8.55/30 | 12/30 | 50% Morfosintaxis 80% Léxico |

Tabla 6: Relación de puntuación, aciertos y fallos de los modelos para chileno.

### 3.5 Español peninsular

De todas las variedades, el español peninsular es el que mejores resultados ha obtenido en la prueba. Así, aunque hay dos modelos (Llama-3.2-3B-Instruct y Yi-1.5.9B-Chat) que obtienen puntuaciones por debajo del aprobado, todos ellos han acertado más de la mitad de las preguntas, como se observa en la Tabla 7.

Los mejores modelos son los de OpenAI (GPT-4o y GPT-4o-mini) y el Gemma, fundamentalmente en las preguntas de tipo léxico, puesto que solo fallan hasta un 10% de estas (P. 30 «¿Qué palabra usas para referirte a un amigo de forma familiar?» R: pana/compa en vez de *colega*). A medida que los modelos van empeorando, los fallos relativos a la morfosintaxis se estabilizan y los fallos de léxico van aumentando paulatinamente, hasta alcanzar un 80% de fallos en el caso de Llama-3.2-3B-Instruct, que ya no identifica léxico propio de la vestimenta o la alimentación, entre otros. Resulta característico que el 88% de los modelos falla en identificar la variante correcta *Esta agua es potable* en vez de señalar la respuesta esperada, por ser la común y característicamente utilizada en esta variedad de español peninsular, *Este agua es potable*.

| Modelos | Puntuación | Aciertos | Frecuencia relativa de fallos |
|---|---|---|---|
| GPT-4o | 28.5/30 | 29/30 | 5% Morfosintaxis 0% Léxico |
| GPT-4o-mini | 27.4/30 | 28/30 | 5% Morfosintaxis 10% Léxico |
| Gemma-2-9b-it | 23.3/30 | 25/30 | 20% Morfosintaxis 10% Léxico |
| Mistral-7b | 22.2/30 | 24/30 | 15% Morfosintaxis 30% Léxico |
| Llama-3.2-11B-Vision-Instruct | 19.4/30 | 22/30 | 20% Morfosintaxis 40% Léxico |
| Occiglot-7b-es-en-instruct | 18.8/30 | 22/30 | 25% Morfosintaxis 30% Léxico |
| Llama-3.1-8B-Instruct | 15.5/30 | 19/30 | 25% Morfosintaxis 50% Léxico |
| Llama-3.2-3B-Instruct | 14.4/30 | 18/30 | 20% Morfosintaxis 80% Léxico |
| Yi-1.5-9B-Chat | 12.5/30 | 17/30 | 40% Morfosintaxis 50% Léxico |

Tabla 7: Relación de puntuación, aciertos y fallos de los modelos para español peninsular.

### 3.6 México y Centroamérica

En español de México y Centroamérica, los modelos tuvieron un rendimiento bajo, como se ve en la Tabla 8. Esto resulta sorprendente, puesto que el español mexicano es la variedad con mayor número de hablantes. Cuatro de los nueve modelos fallan proporcionalmente mucho más en léxico que en morfosintaxis, mientras que el resto tienen porcentajes de error similares para cada tipo.

La pregunta de morfosintaxis más fallada es la número 16, en la que la mayoría de modelos no fueron capaces de identificar «antes de salir yo de la casa» como más típica de México y Centroamérica que «antes de salir de casa». De léxico, la pregunta menos acertada es «El vehículo capaz de transportar a muchas personas es un…» (en esta región, «colectivo» o «bus»).

| Modelos | Puntuación | Aciertos | Frecuencia relativa de fallos |
|---|---|---|---|
| GPT-4o | 17.5/30 | 21/30 | 30% Morfosintaxis 30% Léxico |
| GPT-4o-mini | 17.4/30 | 21/30 | 25% Morfosintaxis 40% Léxico |
| Gemma-2-9b-it | 9.1/30 | 15/30 | 50% Morfosintaxis 50% Léxico |
| Mistral-7b | 6.5/30 | 13/30 | 45% Morfosintaxis 70% Léxico |
| Llama-3.2-11B-Vision-Instruct | 8.8/30 | 15/30 | 45% Morfosintaxis 50% Léxico |
| Occiglot-7b-es-en-instruct | 4.7/30 | 12/30 | 60% Morfosintaxis 60% Léxico |
| Llama-3.1-8B-Instruct | 5.9/30 | 13/30 | 55% Morfosintaxis 60% Léxico |
| Llama-3.2-3B-Instruct | 6.5/30 | 13/30 | 45% Morfosintaxis 80% Léxico |
| Yi-1.5-9B-Chat | 0.9/30 | 10/30 | 55% Morfosintaxis 90% Léxico |

Tabla 8: Relación de puntuación, aciertos y fallos de los modelos para México y Centroamérica.

### 3.7 Rioplatense

La variedad rioplatense es la que más varía en resultados por modelos. Por un lado, es la variedad con el segundo mejor rendimiento para GPT-4o, dos de los modelos de Llama y Gemma. Por otro lado, muestra el peor índice de acierto con Mistral y Occiglot. De estos dos últimos, las preguntas de léxico fueron respondidas de forma incorrecta en un 80% y 90% respectivamente. En general, la mayoría de los modelos fueron capaces de identificar el voseo, el fenómeno morfosintáctico más característico del rioplatense (Alvar, 1996; Palacios 2016; Moreno Fernández, 2014). Los resultados se resumen en la Tabla 9.

| Modelos | Puntuación | Aciertos | Frecuencia relativa de fallos |
|---|---|---|---|
| GPT-4o | 27.1/30 | 28/30 | 10% Morfosintaxis 0% Léxico |
| GPT-4o-mini | 21.4/30 | 24/30 | 25% Morfosintaxis 10% Léxico |
| Gemma-2-9b-it | 16,7/30 | 20/30 | 25% Morfosintaxis 50% Léxico |
| Mistral-7b | 2.5/30 | 10/30 | 60% Morfosintaxis 80% Léxico |
| Llama-3.2-11B-Vision-Instruct | 8.1/30 | 10/30 | 65% Morfosintaxis 70% Léxico |
| Occiglot-7b-es-en-instruct | 2.9/30 | 10/30 | 55% Morfosintaxis 90% Léxico |
| Llama-3.1-8B-Instruct | 9.6/30 | 15/30 | 30% Morfosintaxis 90% Léxico |
| Llama-3.2-3B-Instruct | 12/30 | 14/30 | 35% Morfosintaxis 90% Léxico |

| | 5.4/30 | 12/30 | 60% Morfosintaxis |
|---|---|---|---|
| **Yi-1.5-9B-Chat** | | | 60% Léxico |

Tabla 9: Relación de puntuación, aciertos y fallos de los modelos para la variedad rioplatense.

## 4 Discusión de los resultados

La Figura 1 resume los resultados en forma de mapa de calor de las puntuaciones de los modelos por cada variedad. En ella se puede ver una marcada tendencia: los modelos presentan un desempeño superior en la variedad de español peninsular, mientras que muestran dificultades en la identificación de las características particulares del resto de variedades. En efecto, la española peninsular fue la variedad mejor reconocida por todos los modelos, sobre todo por los de GPT y por Gemma-2-9b-it. La Figura 2 muestra los resultados por categorías y se puede concluir que los errores más frecuentes se dieron fundamentalmente en las preguntas de léxico, especialmente en modelos con menor capacidad. Por otro lado, los modelos, en general, no pudieron reconocer rasgos característicos de las variedades andina, antillana y caribeña continental, pues se observan fallos relativos a fenómenos sintácticos como el loísmo andino y el orden de los pronombres en español antillano y caribeño; y a la identificación de léxico propio de cada variedad. A pesar de que el español mexicano y centroamericano es unas de las variedades con más hablantes, se ha observado una puntuación muy baja, pues los modelos muestran dificultades tanto en la identificación del léxico como de las estructuras sintácticas propias. Los modelos de GPT destacaron frente a otros en el examen de esta variedad, aunque con gran margen de mejora. Estos mismos modelos fueron los únicos que lograron una puntuación alta para las variedades chilena y rioplatense. Aunque fueron capaces de identificar el característico voseo en el español rioplatense, no reconocieron el léxico característico de ninguna de estas dos variedades. En general, los modelos de GPT (GPT-4o y GPT-4o-mini) fueron los mejores en el reconocimiento de las variedades de español, con un desempeño sobresaliente en la variedad peninsular, chilena y rioplatense. Gemma-2-9b-it muestra, por su parte, un buen rendimiento en español peninsular, pero grandes deficiencias a la hora de identificar otras variedades. Se observan porcentajes altos de error en los modelos Mistral-7b-Instruct y los de Llama, por lo que resultan mediocres en esta tarea de identificación. Por último, el modelo Yi-1.5-9B-Chat fue el que obtuvo peores resultados, con fallos considerables en todas las variedades analizadas.

Los resultados confirman, por tanto, que los modelos de lenguaje aún tienen grandes dificultades a la hora de representar la diversidad y riqueza del español. Se observa aún un sesgo lingüístico hacia la variedad peninsular, que puede venir de los propios datos de entrenamiento, y que incapacita a los modelos para responder de forma natural en otras variedades del español. En efecto, al comparar en la Figura 3 la puntuación media de cada variedad con el número total de palabras presentes en CEREAL correspondientes a los mismos países, se observa una fuerte relación que se plasma en un alto coeficiente de correlación de 0.94 entre las dos variables.

Por último, es importante discutir las limitaciones de nuestro estudio. Por una parte, el número de preguntas es reducido lo que aumenta la variabilidad de los resultados. Por otra, pese al cuidado diseño y revisión de las preguntas, puede que alguna no sea representativa de las diferentes variedades o que el conjunto no capture adecuadamente las principales características de las diferentes variedades. Por lo tanto, es aconsejable el desarrollo de tests alternativos y complementarios para las variedades del español con los que se puedan comparar resultados.

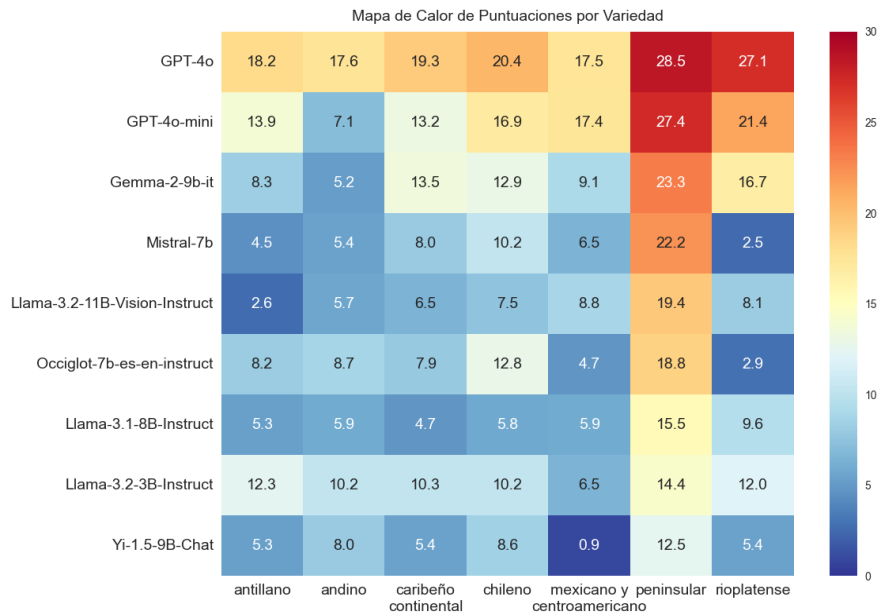

Figura 1: Mapa calor de las puntuaciones de los modelos por cada variedad

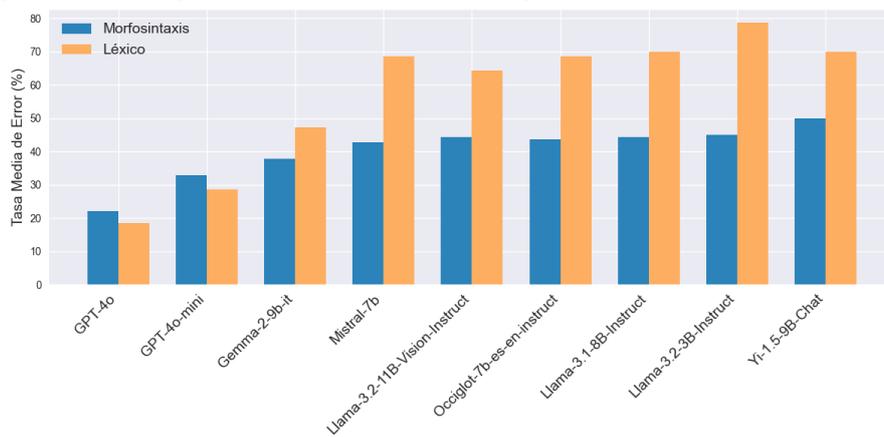

Figura 2: Gráfico de barras de la tasa media de errores de morfosintaxis y léxico por modelo.

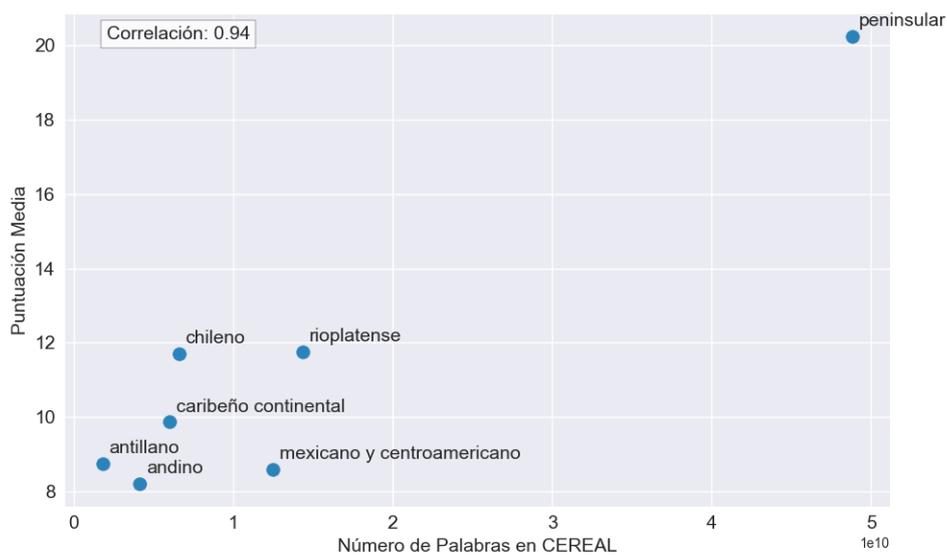

Figura 3: Puntuación media de los modelos en cada variedad versus el número de palabras de cada variedad en CEREAL.

## 5 Conclusiones y líneas futuras

Los resultados del análisis evidencian una gran variabilidad entre las distintas variedades del español y los modelos. Los modelos presentan un desempeño notablemente superior en la variedad de español peninsular, lo que puede sugerir que los datos de entrenamiento estén sesgados hacia esta variedad. Aun así, los errores léxicos de esta variedad aumentan en aquellos modelos de menor capacidad, como los modelos de Llama o el de Yi-1.5-9b-Chat. Por otro lado, los modelos son incapaces de reconocer fenómenos característicos gramaticales de las variedades andina, antillana o caribeña continental. A pesar de que el español mexicano es una de las variedades con mayor número de hablantes, los modelos obtienen puntuaciones sorprendentemente bajas, pues presentan dificultades en la identificación de estructuras sintácticas y léxicas. Para el español chileno y el rioplatense, destacan los de GPT de entre todos los modelos, por presentar un número muy bajo de fallos tanto en léxico como en morfosintaxis. En conclusión, entre las variedades evaluadas, el español peninsular es la mejor identificada por los modelos de lenguaje y, de entre los modelos evaluados, destacan los de GPT por su capacidad para reconocer rasgos léxicos y morfosintácticos característicos de las diferentes variedades del español. Por todo ello, es esencial seguir evaluando modelos, para discernir entre aquellos que puedan ser representativos de la rica diversidad lingüística del español y aquellos otros que contribuyan a incrementar el Sesgo Lingüístico Digital.

## *Agradecimientos*

Este trabajo ha sido posible en parte gracias a los proyectos FUN4DATE (PID2022-136684OB-C22) y SMARTY (PCI2024-153434) financiados por la Agencia Estatal de Investigación (AEI) (con identificador doi:10.13039/501100011033) y al proyecto Europeo SMARTY (Grant 101140087).

## *Bibliografía*

Aleza Izquierdo M. (2010). «Morfología y sintaxis. Observaciones gramaticales de interés en el español de América» en Aleza Izquierdo M. y Enguita Utrilla, J.M. (coords.), *La lengua española en América: normas y usos actuales*. Valencia: Universitat de València.

Alvar, M. (1996). *Manual de dialectología hispánica: el español de América*. Barcelona: Ariel.

Asociación de Academias de la Lengua Española (2010). *Diccionario de americanismos.*

Avramidis et al (2024). *Occiglot at WMT24: European Open-source Large Language Models evaluated on Translation*. In Proceedings of the Ninth Conference on Machine Translation (pp. 292-298).

Bosque, I., Demonte, V. (1999). *Gramática descriptiva de la lengua española.* Madrid: Espasa.

Davies, M. (n.d.) *Corpus del Español: Web/Dialects*. https://www.corpusdelespanol.org/web-dial/

España-Bonet, C., & Barrón-Cedeño, A. (2024). «Elote, Choclo and Mazorca: on the Varieties of Spanish». Proceedings of NAACL-HLT 2024, 3689-3711

Fernández-Ordóñez, I. (1999). «Leísmo, laísmo y loísmo» en Bosque, I. y Demonte, V. (coords.), *Gramática descriptiva de la lengua española*. Madrid: Espasa, 1317-1398.

Grattafiori, A. et al (2024). *The llama 3 herd of models*. arXiv e-prints, arXiv-2407.

Gutiérrez-Rexach, J. (Ed.). (2016). *Enciclopedia de lingüística hispánica* (Vol. 2). New York: Routledge

Jiang, A. et al. (2023). *Mistral* 7b. arXiv preprint arXiv:2310.06825.

Morales, A. (1999). «Anteposición del sujeto en el español del Caribe». Ortiz López, L. A. (Eds.), *El Caribe hispánico: perspectivas lingüísticas actuales*. Madrid/Frankfurt: Iberoamericana Vervuert, 77-98.

Moreno Fernández, F. (2014). *La lengua española en su geografía: Manual de dialectología hispánica*. Madrid: Arco/Libros.

Muñoz-Basols, J., Palomares Marín, M. del M., & Moreno Fernández, F. (2024). «El Sesgo Lingüístico Digital (SLD) en la inteligencia artificial: implicaciones para los modelos de lenguaje masivos en español». *Lengua Y Sociedad*, *23*(2), 623-647. https://doi.org/10.15381/lengsoc.v23i2.28665

PRESEEA. (2016-). *Corpus del Proyecto para el estudio sociolingüístico del español de España y de América*. Alcalá de Henares: Universidad de Alcalá. https://preseea.uah.es/

Real Academia Española (2014). *Diccionario de la lengua española*, 23.ª edición.

Palacios, A. (2016). «Dialectos del español de América: Chile, Río de la Plata y Paraguay» en Gutierrez-Rexach, J. (ed.), *Encilopedia de lingüística hispánica (vol. 2)*. New York: Routledge.

Real Academia Española. (n.d.) *Corpus diacrónico del español (CORDE)*. https://corpus.rae.es/cordenet.html

Real Academia Española. (n.d.) *Corpus del Español del Siglo XXI (CORPES)*. https://www.rae.es/corpes/

Real Academia Española. (n.d.) *Corpus referencial del español actual (CREA)*. https://www.rae.es/crea-anotado/

Riviere G., et al (2024). *Gemma 2: Improving open language models at a practical size.* arXiv preprint arXiv:2408.00118.

Ueda, H. (n.d.) VARILEX, Variación léxica del español en el mundo. https://h-ueda.sakura.ne.jp/varilex/index.html